\documentclass[conference]{IEEEtran}
\ifCLASSINFOpdf
\else
\fi
\usepackage{float}
\usepackage{graphicx}
\usepackage{booktabs}
\usepackage[colorlinks=black,linkcolor=black,hyperindex,CJKbookmarks]{hyperref}
\usepackage{CJK}
\usepackage{indentfirst}
\usepackage{amsmath}
\usepackage{amsfonts}    
\begin{document}
%
\title{Image Classification with A Deep Network Model based on Compressive Sensing}

\author{\IEEEauthorblockN{Yufei Gan, Tong Zhuo, Chu He}
\IEEEauthorblockA{Electronic  Information  School,  Wuhan  University, Wuhan 430072, China\\ 
Email: ganyufei@whu.edu.cn, zhuotong@whu.edu.cn, chuhe@whu.edu.cn}
}

\maketitle

\begin{abstract}
To simplify the parameter of the deep learning network, a cascaded compressive sensing model ``CSNet'' is implemented for image classification. Firstly, we use cascaded compressive sensing network to learn feature from the data. Secondly, CSNet generates the feature by binary hashing and block-wise histograms. Finally, a linear SVM classifier is used to classify these features. The experiments on the MNIST dataset indicate that higher classification accuracy can be obtained by this algorithm.
\end{abstract}

\IEEEpeerreviewmaketitle

\section{Introduction}
Image classification is one of the most fundamental problems in computer vision and pattern recognition.

Recently, Deep Learning has become popular with both industry and academia. A growing number of deep learning techniques is proposed.  As the development of traditional image feature (i.e. SIFT \cite{lowe2004distinctive}, HOG \cite{dalal2005histograms}), Deep Learning can automatically learn feature from training data. A multi-layer structure can help Deep Learning Network learning more abstract semantics features in higher-layer. 

As usual, the Deep Learning network employs a multi-layers network construction.  Cascaded multi-layers network construction could help higher-level features represent more abstract semantic of the data. 

In recent years, the mainstream deep learning approaches are these three: Convolutional Neural Networks (CNNs) \cite{sermanet2013overfeat}\cite{lecun1998gradient}\cite{krizhevsky2012imagenet}, Deep Belief Networks (DBNs), Stacked Auto-Encoders (SAE). A convolutional deep neural network (CNNs) architecture can be structured into two modules: feature extraction module and classifier module. Further, feature module generally comprises of ``three layers'' -- a convolutional filter bank layer, a nonlinear processing layer, and a feature pooling layer. And the classifier module generally comprises fully-connected hidden layers. While many variations of deep learning networks have been proposed, some researchers begin to pay more attention to the architecture of deep learning. 

An example of such research is PCANet \cite{chan2014pcanet} which is proposed by Yi Ma. PCANet use PCA \cite{jolliffe2005principal} filter to replace the convolution filter and the binary quantization is used to replace ReLU \cite{nair2010rectified} as the nonlinear layer. In the output layer, PCANet use the block-wise histograms of the binary codes to generate the feature, and we also can treat the block-wise histograms as the feature pooling layer.

As the research further develops, researchers find the fact that the convolutional deep neural network (CNNs) has weak classification capacity in high-level layer \cite{sharif2014cnn} when compared to SVM. So SVM has been applied to replace the high-level layer recently. 
However, there are still some problems to be solved. Firstly, Convolutional Neural Networks have too many parameters to set, moreover, the performance of the network depends heavily on the setting of the parameter. Secondly, there is not a specific method to classify the high signal to noise ratio images.
In order to solve these problems, we propose the CSNet, which employs compressive sensing technique to deep learning network.
\section{Network}
\begin{figure*}[!bpht]
\centering
\includegraphics[width=7in]{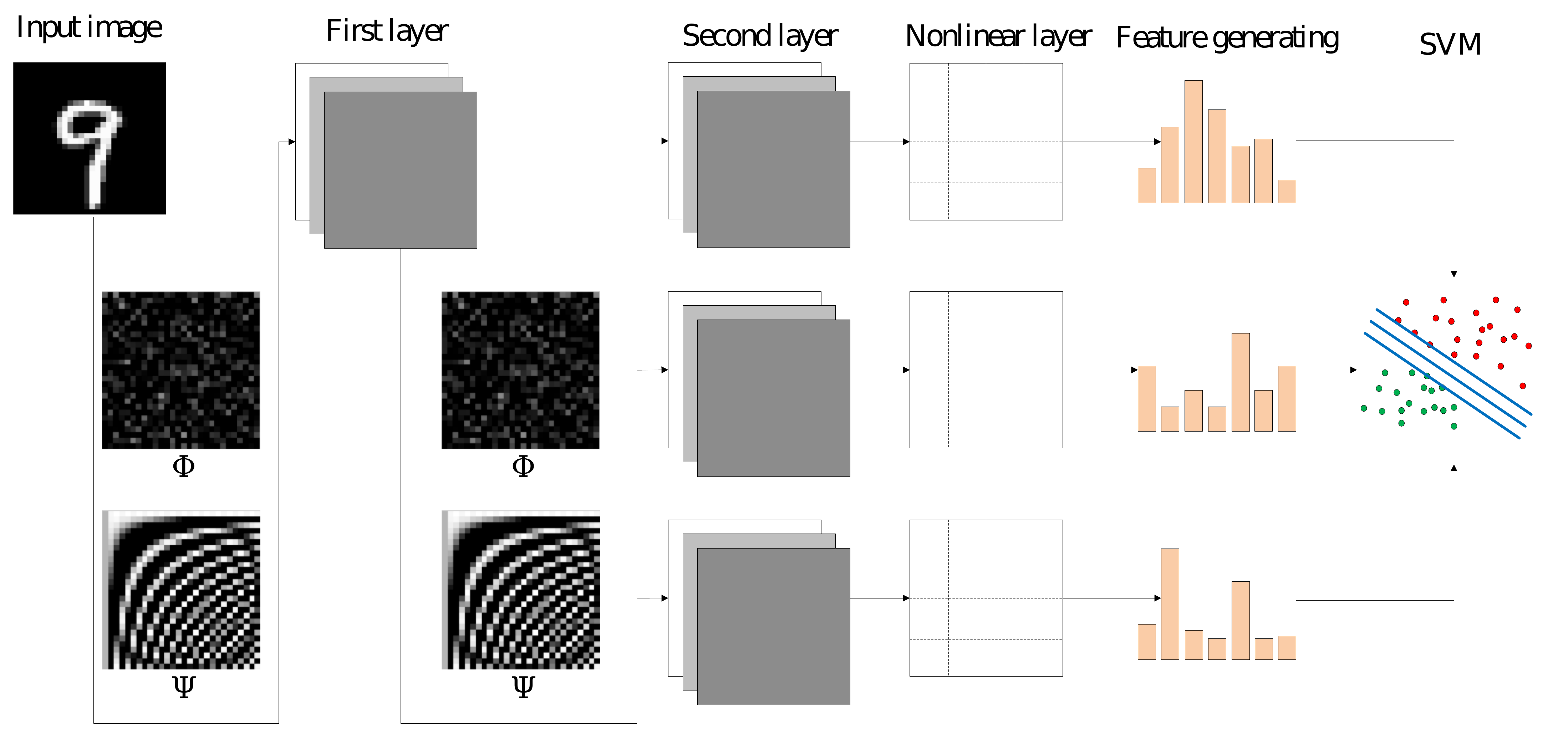}
\caption{The structure of two layers CSNet, the first layer and the second layer are same. The first layer gets $L_1$ maps and the second layer gets $L_1\times{L_2}$ map. In nonlinear layer, binarization operation is applied to  reduce dimension, then use block-wise histogram to generate the feature.} \label{figure:structure}
\end{figure*}
In our CSNet, We use cascaded compressive sensing based on OMP (Orthogonal Matching Pursuit) algorithm \cite{tropp2007signal}\cite{baraniuk2007compressive} to structure multi-level feature learning network, followed by binary hashing operation as a no-linear layer, and use block histograms to output a feature representation of CSNet. This structure is similar to PCANet.
\subsection{Compressive Sensing Algorithm}
Compressive sensing generally comprises of ``three-stage'': getting the sparse representation of signals, computing the measurement of the data, recovering the data from the measurement. Recovering algorithms can be concluded with a minimization problem. As usual, we apply these two kinds of methods to solve the minimization problem: greedy method and Convex Optimization Methods.

We select DCT transform to sparse the image, and we use random gauss matrices to compute the measurement of the data. Considering the training efficiency, we employ the OMP algorithm to recover the data since the greedy algorithm has low complexity. By this way, we can reach the balance of the training speed and training efficient.
\subsection{Orthogonal Matching Pursuit algorithm in CSNet}
We suppose that there are $N$ input image $\{I_i\}^N_{i=1}$, each image has the same size $m\times{n}$,  and assume that the patch size is  $k_1\times{k_2}$ at all stages. In our CSNet, we assume that the number of filters in layer \emph{i} is $L_{i}$ . We denote \emph{i}th image by $X_i=\left[x_{i,1},x_{i,2},\cdots,x_{i,mn}\right]$ where each $x_{i,j}$ denotes the \emph{j}th vectorized patch in $I_{i}$. Then we subtract patch mean from each patch and obtain mean-removed patch $
\bar{X}_i=\left[\bar{x}_{i,1},\bar{x}_{i,2},\cdots,\bar{x}_{i,mn}\right]$. Finally, we putting all image together:
\begin{equation}
X=\left[\bar{x}_{i,1},\bar{x}_{i,2},\cdots,\bar{x}_{i,mn}\right]\in\mathbb{R}^{k_1k_2\times{Nmn}}.\end{equation}

Now, we begin to introduce the core algorithm of CSNet:

Firstly, we process data with (random Gaussian) measurement matrix $\Phi$ and discrete cosine transform (DCT) matrix $\Psi$.  The process can be summarized as follow equation:
\begin{equation}Y=\Phi\Psi XX^T\end{equation}

We initialize the residual $r_0=y$, and define measurement matrixs columns by $\varphi_1,\varphi_2,\cdots,\varphi_d$. For each row of $Y$, we find the index $\lambda_t$ that solves the easy optimization problem
\begin{equation}
\lambda_t={\arg\max}_{j=1,\cdots,N}|\langle r_{i-1},\varphi_t\rangle|
\end{equation}

Secondly, update the index set and the matrix of chosen atoms:
\begin{equation}
\Lambda_t=\Lambda_{t-1}\cup\Lambda_t
\end{equation}
\begin{equation}
\Phi_t=\left[\Phi_{t-1},\varphi_t\right],
\end{equation}

Thirdly, solve a least squares problem, then get a filter parameter and save in $W$, and the $s$ represents the row number of signal $X$.
\begin{equation}
\tilde{x}_t=\arg\min\|y-\Phi_t\tilde{x}\|_2
\end{equation}
\begin{equation}
W_{s,t}=f(\lambda_t,x_t)
\end{equation}
The function means that the value of $\hat{s}$ in component $\lambda_i$ equals the $j$th component of $\tilde{x}_t$.

Finally, calculate the new approximation of the data and the new residual:
\begin{equation}
r_t=y-\phi_t\tilde{x}_t
\end{equation}

We repeat above three stage $K$ times with increasing $t$, and $K$ could be treated as sparsity level. And the filters of CSNet can be expressed as the recovery of $W_{s,t}$
\begin{equation}
 W_l=[W_{1,l};W_{2,l};\cdots;W_{k_1,l}]\in\mathbb{R}^{k_1\times k_2},l=1,2,\cdots,L_1
 \end{equation}

\subsection{Cascaded Compressive Sensing Network}
Let the $l$th filter output of the first layer be:
\begin{equation}\label{equ:firstlayer}
I^l_i=I_i\ast W^k_l,i=1,2,\cdots,N, 
\end{equation}
The $\ast$ operation denotes 2D convolution operation, and every compressive sensing layer is same as the first compressive sensing layer. Assume the CSNet have $c$ compressive sensing layers the last compressive sensing output be:
\begin{equation} \label{equ:secondlayer}
O^l_i=I_i\ast W^c_l  
\end{equation}
The introduction given above has concluded all the core algorithm, and the first layer and the second layer in Figure \ref{figure:structure} illustrate the process  (\ref{equ:firstlayer}) or (\ref{equ:secondlayer}).

\section{Image Classification Based On CSNET}

\subsection{Generating Feature}
In non-linear layer, we apply the simplest non-linear operation -- Heaviside function:
\begin{equation}
H(x)=\left\{
\begin{aligned}
&1&,&~if~x>0~\\
&0&,&~otherwise.
\end{aligned}
\right.\\
\end{equation}
In order to reduce the dimensionality, we transform a binary number to a decimal number by function $H(I_i\ast W^k_l)$. This process is similar to pooling operation:
\begin{equation}
T^l_i=\sum^{L_2}_{l=1}2^{l-1}H(I_i\ast W^k_l),
\end{equation}
We use block-wise histograms to generate the feature, and the local block can be either overlapping or non-overlapping.
\begin{equation}
f_i=\left[Bhist(T^1_i),Bhist(T^2_i),\cdots,Bhist(T^{L_i}_i)\right]^T\in\mathbb{R}^{(2^{L_2})L_1B}
\end{equation}
Now, we get the feature for each image. We can control the feature dimensionality by set the number of the filters.
\subsection{Train and Test}
The structure of CSNet is shown in Figure \ref{figure:structure}. In this figure, CSNet has two compressive sensing layers. We treat the network which comprises cascaded compressive sensing, binary quantization and block-wise histograms as a feature extractor. So we use libsvm \cite{chang2011libsvm} with trade-off parameter set to $C=1$. And when we train the CSNet, the filters can be computed and the parameter of SVM will be trained. Once the filters and SVM are determined, CSNet can be applied to classify image. 
\begin{figure*}[!ht]
\centering
\includegraphics[width=7in]{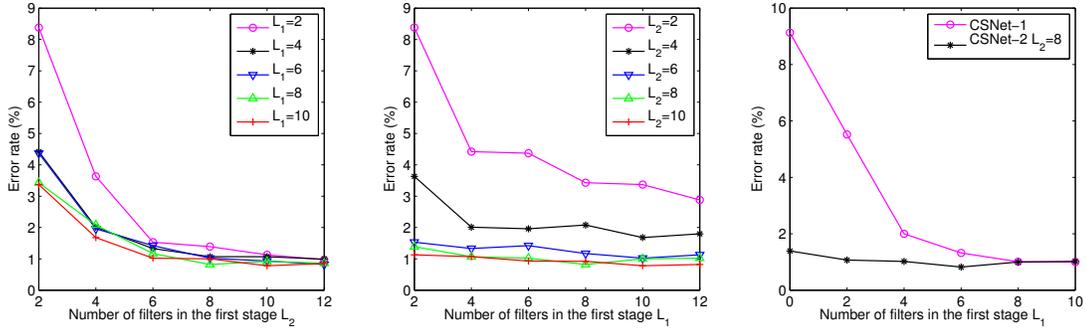}
\caption{Left figure and middle figure: the impact of the number of filters. Right figure: the impact of the number of layer}\label{figure:3}
\end{figure*}
\section{Experiment}
Experiments are conducted on the MNIST dataset, this dataset have 60000 images for training and 10000 images for test, and all the images are of size $28\times28$. In order to compare with PCANet, we use the subset of MNIST which is given by the demo of PCANet, in the demo of PCANet, 12000 images are given to train the network, and 50000 images for test, we use these 50000 images in the demo of PCANet to train network, and others for test. We use two layers CSNet and one layer CSNet to test the classification performance of CSNet, and use PCANet for comparison, And the Table \ref{table:noise} show the best performance of the PCANet and CSNet.

\subsection{Impact of the parameter}
\subsubsection{Impact of the number of filters}
We vary the number of filters in the first stage $L_1$ (from 2 to 12) and the second stage $L_2$ (from 2 to 12). The overlapping rate is set as 0, the filter size of the network is 7 and 7 ($k_1=k_2=7$), the block size is 7 and 7. We use 50000 images to train CSNet, and 12000 images for test. The results are shown in Figure \ref{figure:3} (left figure and middle figure). In this figure, we can find we can improve the accuracy of classification by increase the number of the filters, and the number of the filters in the second layers can enhance the insufficient number of the filters in first layers. Inversely, if the number of the filters in second is insufficient, it is hard to enhance the performance of the classification by increasing the number of the filters in the first layer.

\subsubsection{Impact of the number of layers}
To explore the performance difference of multi-layer CSNet with single-layer CSNet. In this experiment, some parameter are same ($PatchSize=7, BlkOverLapRatio=0, scale=1, HistBlockSize_1=7, HistBlockSize_2=7$). We conduct the experiment on the single-layer and multi-layer ($L_2=8$), in this experiment we also use 50000 images to train CSNet, and 12000 images for test, The overlapping rate is set as 0, the filter size of the network is 7 and 7 ($k_1=k_2=7$), the block size is 7 and 7. The results are shown in Figure \ref{figure:3} (Right figure). From this picture, we can confirm a fact that two-layer CSNet ($L_2=8$) can get the lower error rate the single-layer CSNet in same number of filter in the first layer. And we can also find the fact that the difference between two-layer CSNet ($L_2=8$) and single-layer CSNet becomes narrow when the number of the filers in first layer reach 8. It may be due to the data is relatively simple.
\subsection{Impact of the noise }
\begin{figure}[!ht]
\centering
\includegraphics[width=3.5in]{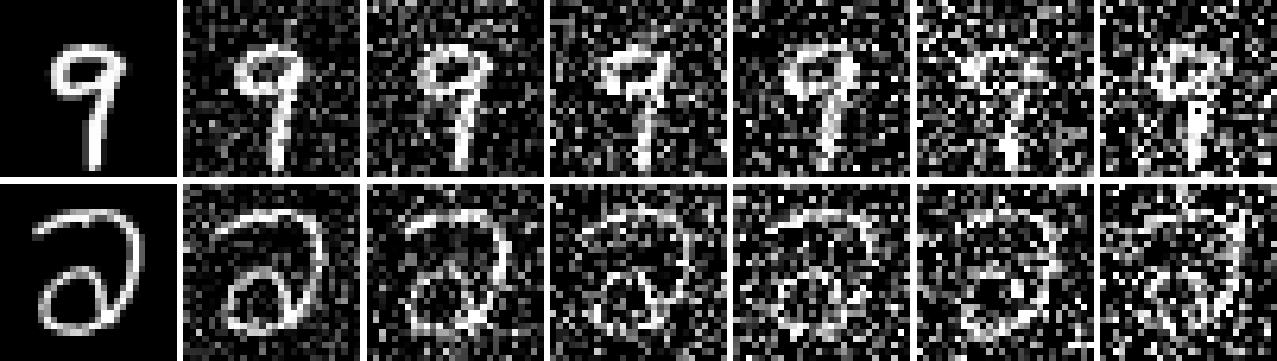}
\caption{Impact of the noise, Gaussian noise is added to each train image and test image.}\label{figure:noise}
\end{figure}

\begin{table}[!ht]\
\caption{}\label{table:noise}
\centering
the error rate of image classification with noise \\
~\\
\centering
\begin{tabular}{cccccccc} \toprule 
Variance	&   0		&	0.05	&	0.10 	& 0.15 	& 0.20	& 0.25    	& 0.30 \\ \midrule
CSNet	& 0.8\%	&2.7\%	& 4.97\%	& 7.37\%	&9.76\%	&14.96\% 	&15.7\%  \\  \bottomrule
\end{tabular}\\
~\\
Note that the best performance of CSNet is error rate \textbf{0.8\%}, while the best performance of PCANet is error rate \textbf{1.0\%} in our experiment.
\end{table}

We test CSNet in different SNR (signal to noise rate). Gaussian noise is added to each train image and test image. Figure \ref{figure:noise} show the processed data, in this picture, the mean of Gaussian noise is set to zero and variance from 0 to 0.30. Actually, when variance equals to 3, the digital has become illegible.  We use two layers CSNet ($L_1=8, L_2=8, PatchSize=7, BlkOverLapRatio=0, scale=1, HistBlockSize_1=7, HistBlockSize_2=7$) to The experimental results are given in Table \ref{table:noise}.  Although the images are difficult to identify, CSNet has good performance (error rate when variance is 0.25)
\subsection{Visualize the learned CSNet}
\begin{figure}[!ht]
\centering
\includegraphics[width=3.5in]{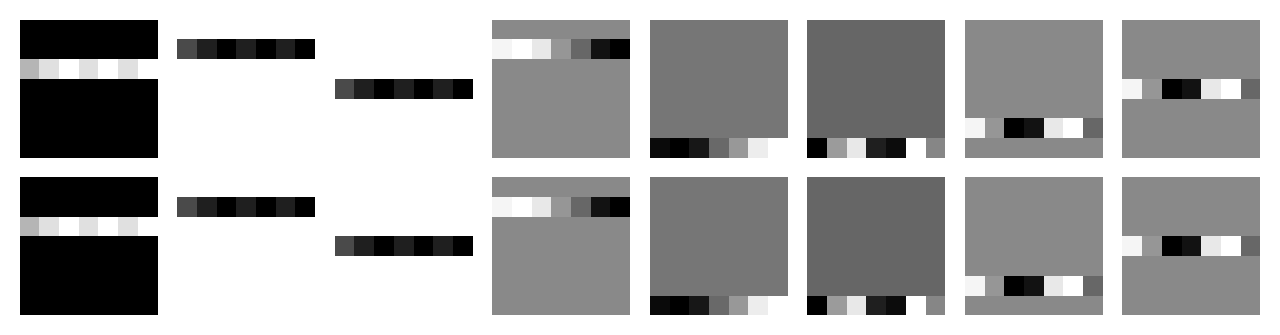}
\caption{The filters learned from CSNet (the filter has been multiplied by DCT transformation matrix) on MNIST. Top row: the first stage. Bottom row: the second stage.}\label{figure:CSNet}
\end{figure}
\begin{figure}[!ht]
\centering
\includegraphics[width=3.5in]{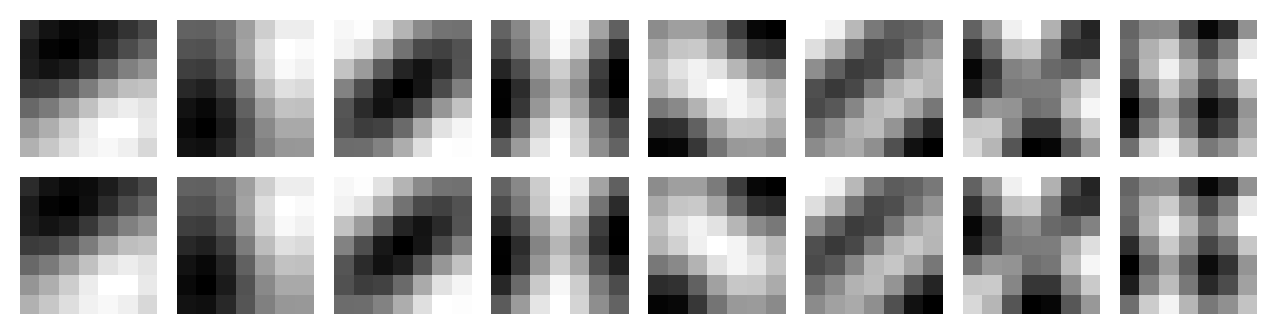}
\caption{The filters learned from PCANet on MNIST. Top row: the first stage. Bottom row: the second stage.}\label{figure:PCANet}
\end{figure}
\begin{figure}[!ht]
\centering
\includegraphics[width=3.5in]{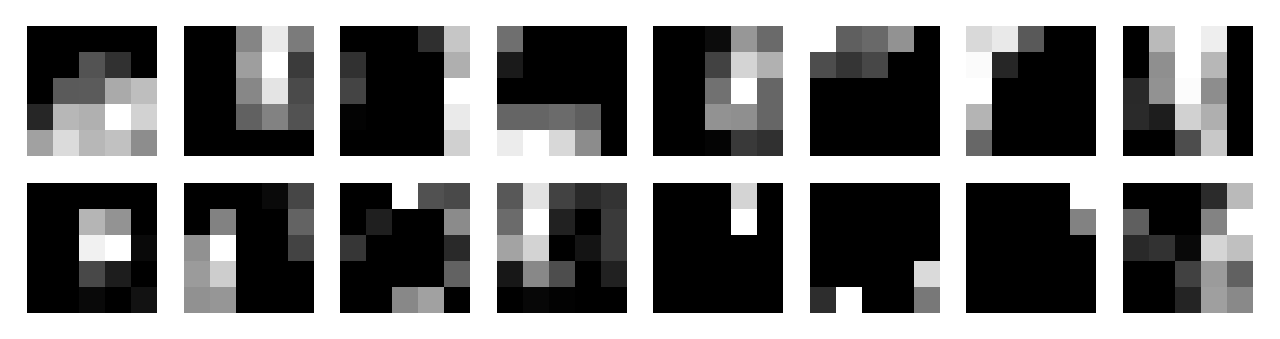}
\caption{The filters learned from CNNs on MNIST. Top row: the first stage. Bottom row: the second stage.}\label{figure:CNNet}
\end{figure}
We draw the learned CSNet filters (the filter has been multiplied by DCT transformation matrix) in Figure \ref{figure:CSNet}. The filters show a characteristic of the random sampling. According to the position of white point in filter, the filters actually can be treated as different frequency filters. In the figure, the left filters are low-pass filters, and the right filters are high-pass filters. To compare our CSNet, we draw the learned PCANet filters in Figure \ref{figure:PCANet} and the learned CNNs filters in Figure \ref{figure:CNNet}. And the size of filter in PCANet is $7\times7$, the size of filter in CNNs is $7\times7$, the learned CNNs filters from MNIST is different from \cite{zeiler2011adaptive}, it may be caused by the insufficiency of training epoch and the difference between the data set. We use the same 50000 training data and 12000 test data to train and test these networks. In the figure, the filter of PCANet and CNNs show an basic feature (i.e. edges and blobs), and CSNet filters which has been multiplied by DCT transformation matrix are similar to sample matrix.
\section{Conclusion}
In this paper, we proposed a deep learning network based on the compressive sensing. CSNet use compressive sensing algorithm to as main feature learning layer, then get the feature representation of input images by binary hashing and block histogram. Using CSNet to compute filter does not require numerical optimization solver so the training process can be extremely efficient. CSNet inherits noise immunity of compressive sensing thanks to cascaded compressive sensing structure of CSNet.

Our results indicate that CSNet can perform fast and accuracy in MNIST datasets. However, MNIST dataset still has high SNR, it can be thought caused by the fact that specific of input images is not distinct enough. But it is too rigid to obtain a high distinct dataset. Anther conclusion is that effect of insufficient low-level features is difficult to improve by increasing the number of the semantic features (the number of filters in second layer). The experiment about the difference between two-layer CSNet and single-layer CSNet indicate the fact that the multi-layer network could contribute the accurate rate, and the deep network structure might the key reason to develop the performance of image classification.

In feature work, we hope to apply more efficient compressive sensing recovery algorithms to CSNet, thus Our CSNet can train faster. And the experiments will be conducted in more datasets.

\section*{Acknowledgement}
The work was supported by the National Key Basic Research and Development Program of China (973 Program) (No.2013CB733404), NSFC grant (No.41371342, No.61331016) and the China Postdoctoral Science Foundation funded project and the Natural Science Foundation of Hubei Province.

\bibliographystyle{IEEEtran}
\bibliography{G}
\end{document}